\documentclass[10pt,twocolumn,letterpaper]{article}
\usepackage[pagenumbers]{cvpr} 

\newcommand{\red}[1]{{\color{red}#1}}

\definecolor{cvprblue}{rgb}{0.21,0.49,0.74}
\usepackage[pagebackref,breaklinks,colorlinks,allcolors=cvprblue]{hyperref}
\usepackage{url}
\usepackage{booktabs}
\usepackage{graphicx}
\usepackage{multirow}
\usepackage{subcaption}
\usepackage{wrapfig}
\usepackage{adjustbox}
\usepackage{enumitem}
\usepackage{algorithm}
\usepackage{algorithmic}
\usepackage{setspace}
\usepackage{bm}
\usepackage{makecell}
\usepackage[accsupp]{axessibility}

\definecolor{Gray}{rgb}{0.95, 0.95, 0.95}
\newcommand{\blue}{\textcolor{blue}}
\newcommand{\gray}{\textcolor{gray}}

\def\paperID{6393} 
\def\confName{CVPR}
\def\confYear{2025}

\title{Uncertainty-guided Perturbation for Image Super-Resolution Diffusion Model}

\author{Leheng Zhang \quad Weiyi You \quad Kexuan Shi \quad Shuhang Gu\footnotemark[1]\\
University of Electronic Science and Technology of China  \\
{\tt \small \{lehengzhang12, shuhanggu\}@gmail.com} \\
{\small \url{https://github.com/LabShuHangGU/UPSR}}
}

\begin{document}

\maketitle

\renewcommand{\thefootnote}{\fnsymbol{footnote}}
\footnotetext[1]{corresponding author}

\begin{abstract}
Diffusion-based image super-resolution methods have demonstrated significant advantages over GAN-based approaches, particularly in terms of perceptual quality. 
Building upon a lengthy Markov chain, diffusion-based methods possess remarkable modeling capacity, enabling them to achieve outstanding performance in real-world scenarios.
Unlike previous methods that focus on modifying the noise schedule or sampling process to enhance performance, our approach emphasizes the improved utilization of LR information.
We find that different regions of the LR image can be viewed as corresponding to different timesteps in a diffusion process, where flat areas are closer to the target HR distribution but edge and texture regions are farther away. 
In these flat areas, applying a slight noise is more advantageous for the reconstruction.
We associate this characteristic with uncertainty and propose to apply uncertainty estimate to guide region-specific noise level control, a technique we refer to as Uncertainty-guided Noise Weighting.
Pixels with lower uncertainty (i.e., flat regions) receive reduced noise to preserve more LR information, therefore improving performance.
Furthermore, we modify the network architecture of previous methods to develop our \textbf{U}ncertainty-guided \textbf{P}erturbation \textbf{S}uper-\textbf{R}esolution (\textbf{UPSR}) model.
Extensive experimental results demonstrate that, despite reduced model size and training overhead, the proposed UPSR method outperforms current state-of-the-art methods across various datasets, both quantitatively and qualitatively.
\end{abstract}
    
\section{Introduction}
\begin{figure}
    \centering
    \includegraphics[width=\linewidth]{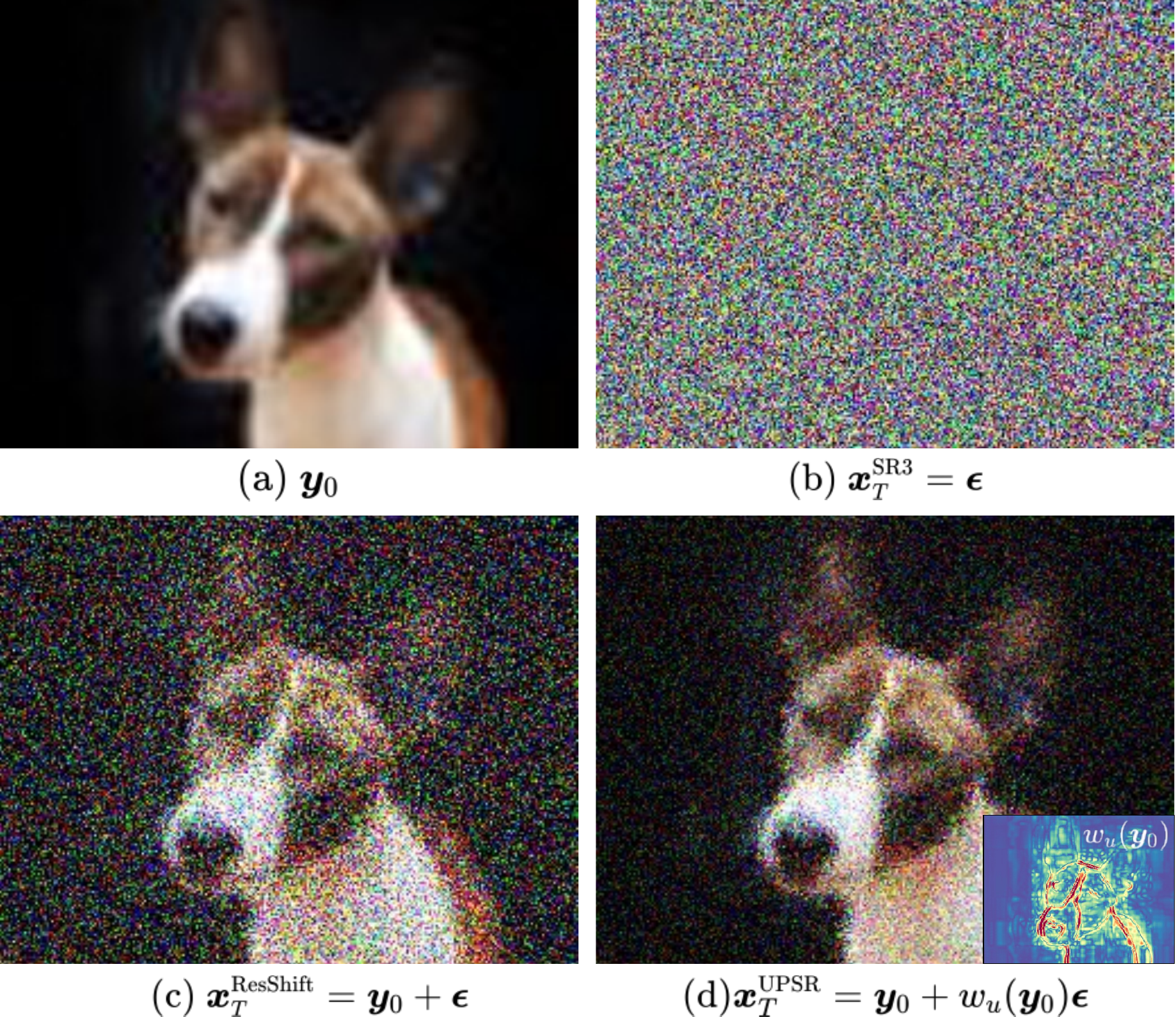}
    
    \caption{A comparison of initial state setup between different diffusion-based image super-resolution methods, where $\bm{\epsilon}\sim \mathcal{N}(\bm{0},\sigma_{\text{max}}^2\bm{I})$. (b) SR3~\cite{saharia2022image} initiate the diffusion process from pure Gaussian noise, whereas (c) ResShift~\cite{yue2024resshift} and (d) our UPSR embed the LR input into the initial noise map. Additionally, we apply uncertainty-guided weighting coefficient $w_u(\bm{y}_0)$ to reduce the noise level in flat areas, achieving a more specialized diffusion process for SR to improve performance.}
    \label{fig: teaser1}
    
\end{figure}
Single image super-resolution (SR), which aims to recover clean the high-resolution (HR) image from its degraded, contaminated low-resolution (LR) counterpart, is a classical problem in the computer vision society.
Significant information loss during the degradation process underscores the need for robust super-resolution modeling capability to recover missing details and produce visually pleasing results.
Existing SR methods have explored a variety of advanced network architectures and complex degradation modeling to improve performance in classical SR~\citep{liang2021swinir, chen2023activating, zhang2024transcending} and real-world SR~\citep{zhang2021designing, wang2021real}, respectively.

Recently, diffusion models~\citep{ho2020denoising, song2020denoising, karras2022elucidating} have demonstrated impressive capability in image synthesis, offering a promising new approach for the real-world image super-resolution task.
These methods transform pure Gaussian noise into high-quality images through a predefined Markov chain, and their solid theoretical foundation endows them with exceptional modeling capacity to bridge diverse data distributions.
To leverage the modeling capabilities of diffusion models for recovering missing details in LR images, several methods~\citep{saharia2022image, rombach2022high} begin the SR process by sampling from a standard Gaussian distribution and gradually refine the noisy inputs into high-quality outcomes.
%
However, starting from pure noise was originally intended for image synthesis tasks, resulting in suboptimal outcomes for super-resolution. 
Additionally, these methods typically require a lengthy sampling process, limiting their practicality for real-world applications.

%
To address the aforementioned issues, ResShift~\citep{yue2024resshift} focuses on the construction of the prior distribution, highlighting the importance of LR information during the diffusion process.
By embedding the LR image into the initial noise map and progressively recovering the residual between the LR and HR images, ResShift greatly simplifies the diffusion process. 
Instead of modeling the entire HR image from noise, it only needs to estimate the LR-HR residual, which shortens the sampling process while significantly enhancing super-resolution results. 
Despite improved performance, challenges remain, as shown in Fig.~\ref{fig: teaser1} where most details are obscured by heavy noise. 
This still poses additional challenges, as exploiting the information from surrounding areas is crucial for recovering missing details in the SR task. 

For the pursuit of designing a more specialized and effective diffusion process for SR to improve performance, we first propose to make better use of the LR information in the prior distribution.
In recent studies, little attention has been given to the inherent information in the LR image that flat areas are already close to the target, whereas edge and texture regions are farther away.
To leverage this information, we propose to consider different regions in the LR image as being situated at various timesteps of an isotropic diffusion process.
This leads to an anisotropic diffusion process, where flat areas are assigned lower noise levels (as $t\to 0$), while edge and texture regions receive larger noise (as $t\to T$).
To achieve region-specific noise control, we draw inspiration from uncertainty-driven SR approaches~\citep{lee2019gram, ning2021uncertainty} and employ a simple SR network to estimate the uncertainty (variance) of different areas in the input LR image.
This uncertainty reflects the difficulty of recovering HR details and is closely related to the distribution disparity between LR and HR, indicating the amount of noise required. 
We then propose a technique called Uncertainty-guided Noise Weighting, which applies weight coefficients to noise levels in different regions based on their uncertainty estimates. 
The weight coefficient is positively correlated with the uncertainty estimate, enabling adaptive reduction of noise intensity in flat areas to preserve more details in the initial state of the diffusion process.
Equipped with UNW and a modified network architecture, we introduce the UPSR model, which achieves state-of-the-art performance while reducing computational costs, as validated on both synthetic and real-world SR datasets.

Our major contributions are summarized as follows:
\begin{itemize}[noitemsep,topsep=0pt,leftmargin=*]
    \item 
    To achieve a more specialized diffusion pipeline for SR, we propose to adjust the noise level for different areas based on their uncertainty estimate, where the noise in flat areas (low uncertainty) is reduced. This Uncertainty-guided Noise Weighting strategy enables adaptive control across different areas and results in improved performance.
    \item We establish the connection between the residual estimated by a pre-trained SR network and the uncertainty of the LR input. The estimated residual could serve as an approximate measure of uncertainty, enabling effective uncertainty estimation in the proposed uncertainty-guided noise weighting scheme.
    \item By integrating the proposed uncertainty-guided noise weighting scheme and the modified network architecture, our method achieves state-of-the-art performance across various benchmark datasets with smaller model size and less training overhead.
\end{itemize}

\section{Related Works}
\label{related_works}

\subsection{Image Super-Resolution}
The current literature on image super-resolution can be divided into two categories: classical SR and real SR.
The former focuses on addressing predefined degradation patterns, e.g., bicubic downsampling.
After the era of conventional methods~\cite{yang2010image, zeyde2012single, gu2015convolutional}, a multitude of works~\citep{dong2015image, lim2017enhanced, zhang2018image, liang2021swinir, chen2023activating, li2023efficient, zhang2024transcending} focus on exploring the potential of neural network, ranging from improved convolutional neural networks to transformer variants~\citep{vaswani2017attention}.
These architectural modifications enhance the capacity of modeling complex relationship between LR and HR images, thereby achieving better performance.
Despite their success, these methods perform poorly when directly applied to the real SR task, where the degradation model includes a series of unknown and complex noise and blur patterns.
BSRGAN~\citep{zhang2021designing} incorporates a series of degradation operators to simulate real-world scenarios, providing a robust training baseline.
By combining this strategy with a generative adversarial network~\citep{goodfellow2014generative}, BSRGAN achieves remarkable fidelity and perceptual quality in the real SR task.
Subsequently, RealESRGAN~\citep{wang2021real} extends this scheme to a high-order degradation model, providing a more robust training data synthesis procedure and better performance.

\subsection{Diffusion Models}
Rooted in nonequilibrium thermodynamics, diffusion probabilistic models~\citep{sohl2015deep} have emerged as a new trend in recent research of generative models.
DDPM~\citep{ho2020denoising} proposes a lengthy parameterized Markov chain to bridge the distribution of high-quality images and the standard Gaussian distribution.
Built upon solid theoretic foundations, DDPM simplifies training objectives and achieves a great breakthrough in image synthesis tasks, demonstrating significant generative capacity compared to GAN-based methods~\citep{goodfellow2014generative, miyato2018spectral, karras2019style}.
Subsequently, immense amounts of related research have developed various techniques to further improve DDPM by enhancing performance or shortening sampling process.
These improvements include deterministic sampling~\citep{song2020denoising}, modified noise schedule~\citep{nichol2021improved}, harnessing latent space~\citep{rombach2022high}, second-order ODE solver~\citep{karras2022elucidating}, and improved training dynamics~\citep{karras2024analyzing}.

\subsection{Diffusion-based Image Super-Resolution}
SRDiff~\citep{li2022srdiff} and SR3~\citep{saharia2022image} are the first to apply diffusion models to image super-resolution by taking the LR input as conditional information, demonstrating the efficacy of diffusion model in generating perceptually high-quality SR images.
Despite superior performance, SR3 suffers from bias issues and a costly sampling process.
LDM-SR~\citep{rombach2022high} trains an autoencoder and performs the diffusion process in its low-dimensional latent space.
This approach allows the model to concentrate on perceptually relevant details and significantly improves computational efficiency.
ResShift~\citep{yue2024resshift} embeds the LR information $\bm{y}_0$ into the initial state $\bm{x}_T$ and built a new prior distribution: $\bm{x}_T \sim \mathcal{N}(\bm{x}_T \mid \bm{y}_0, \kappa^2 \eta_T \bm{I})$. 
In this way, the diffusion process shifts from generating the HR image $\bm{x}_0$ from pure noise $\bm{\epsilon}$ to generating the residual $\bm{x}_0 - \bm{y}_0$ given noisy LR information $\bm{y}_0+\bm{\epsilon}$, greatly reducing the difficulty of estimation.
The forward and backward transition distributions are defined as:
\vspace{-1mm}
\begin{equation}
\label{eq: resshift forward transition}
    \begin{split}
        &q(\bm{x}_t \mid \bm{x}_{t-1}, \bm{x}_{0}, \bm{y}_0) = \\
        &\quad \quad \quad \quad \mathcal{N} \left (\bm{x}_t \mid \bm{x}_{t-1} + \alpha_t (\bm{y}_{0} - \bm{x}_0), \kappa^2 \alpha_t \bm{I} \right ) \\ 
    \end{split}
\vspace{1mm}
\end{equation}
and
\vspace{-1mm}
\begin{equation}
\label{eq: resshift backward transition}
    \begin{split}
        &q(\bm{x}_{t-1} \mid \bm{x}_{t}, \bm{x}_{0}, \bm{y}_0) = \\
        &\quad \quad \quad \quad \mathcal{N} \left ( \bm{x}_{t-1} \mid \frac{\eta_{t-1}}{\eta_t} \bm{x}_{t} + \frac{\alpha_t}{\eta_t} \bm{x}_0, \kappa^2 \frac{\eta_{t-1}}{\eta_t} \alpha_t \bm{I} \right ), 
    \end{split}
\vspace{-1mm}
\end{equation}
for $t=1, 2, \cdots, T$; $\eta_t$ and $\alpha_t = \eta_t - \eta_{t-1}$ are time-dependent positive parameters that characterize the velocity of mean shift and noise injection at different timesteps.
In this work, we adopt the sampling process and network architecture of ResShift as our baseline and propose a series of modifications to the sampling pipeline and network architecture to enhance performance while significantly saving model size and training overhead.

\section{Methodology}

\begin{figure}[tbp]
    \centering
    \vspace{1.5mm}
    \begin{subfigure}[b]{0.65\linewidth}
        \centering
        \includegraphics[width=\linewidth]{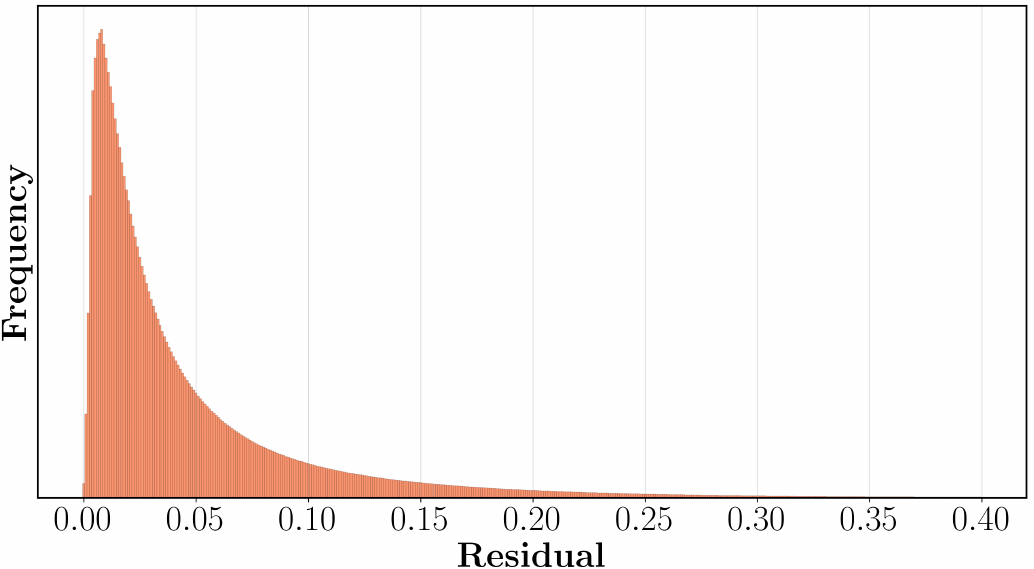}  
        \caption{}
        \label{fig: distribution}
    \end{subfigure}
    \vfill
    \begin{subfigure}[b]{\linewidth}
        \centering
        \includegraphics[width=\linewidth]{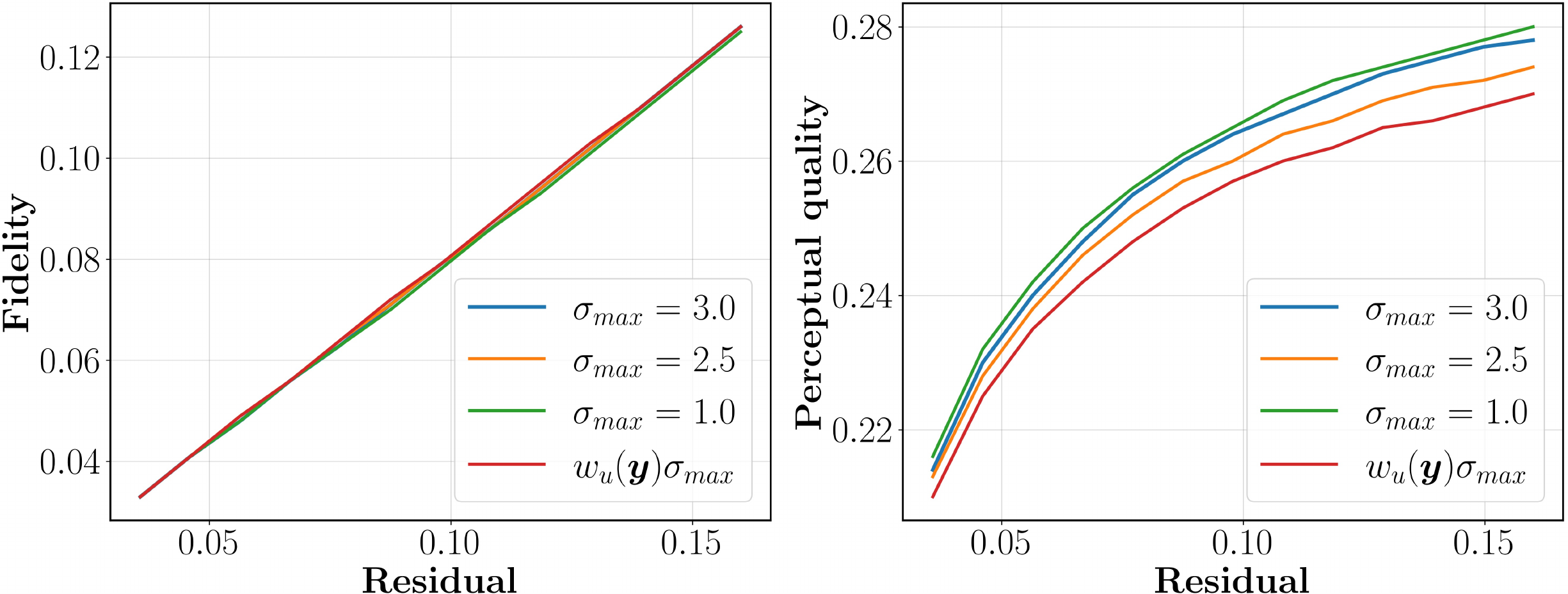}  
        \caption{}
        \label{fig: res}
    \end{subfigure}
    
    \vspace{-3mm}
    \caption{\textbf{(a)} The distribution of pixel \textbf{residual} $|y - x|$ computed on ImageNet-Test dataset~\cite{yue2024resshift}, omitting values where $|y - x| > 0.4$ for clarity. The result exhibits a distinct long-tailed characteristic. \textbf{(b)} The statistical curves of \textbf{fidelity} $|f(y) - x|$ and \textbf{perceptual quality} $|\phi(f(y))-\phi(x)|$ with respect to \textbf{residual} $|y - x|$ under different noise levels. 
    As $|y - x|$ increases, the gap of fidelity remains relatively stable when different noise levels are applied.
    In contrast, the perceptual quality is more sensitive to the noise level.
    A larger noise is more requisite in regions with high residual value to achieve better perceptual quality.
    Meanwhile, we propose weighted noise level $w_u(\bm{y})\sigma_{max}$ which could lead to better results, with details presented in Sec.~\ref{sec: uncertainty-based noise}.
    }
    \label{fig:layout}
    \vspace{-2mm}
\end{figure}

\subsection{Noise Levels in Diffusion-based SR}
\label{sec: influence}
    
    

In the context of SDE~\cite{song2019generative,song2020score}, perturbing the data with Gaussian noise is crucial for facilitating score estimation~\cite{hyvarinen2005estimation}, especially for data residing in low-dimentional manifolds.
They apply various noise levels to perturb the data to explore the low-density regions and enable the score estimation towards high-density regions.
In the super-resolution task, the target is to estimate the gradients from LR images (low density) to HR images (high density) in the real image distribution.
Recent diffusion-based methods~\cite{yue2024resshift, yue2024efficient} inject a large initial noise into the LR image and gradually reduce it throughout the reverse diffusion process, as the LR sample progressively moves closer to high-density regions.
However, these methods focus solely on the different amounts of noise needed at different timesteps during the diffusion process, without accounting for the fact that different areas of each image require varying noise intensities.
To provide a clearer insight, we conduct several experiments to demonstrate how the fidelity and perceptual quality of the output change when different overall noise levels are applied to a diffusion-based SR model.

We first analyze the distribution of the residual $|y - x|$ between each pixel pair of the upsampled LR image $\bm{y}$ and its corresponding HR image $\bm{x}$, as illustrated in Fig.~\ref{fig: distribution}.
The result clearly indicates that $|y-x|$ follows a long-tailed distribution, with over 95\% of the data concentrated in the range $[0.01, 0.16]$.
Next, we focus on data within this range to investigate the relationships between the residual $|y-x|$, fidelity $|f(y)-x|$, perceptual quality $|\phi(f(y)) -\phi(x)|$ of the output of the denoiser $f(\cdot)$ trained under different noise levels $\sigma_{max}$ (isotropic and anisotropic).
The results are illustrated in Fig.~\ref{fig: res}. 
Due to the blurring and downsampling degradation, the residual $|y-x|$ tends to be low in flat areas and high in regions characterized by edges and textures.
As $|y-x|$ grows, the perceptual quality gap between $|\phi(f(y))-\phi(x)|_{\sigma_{max}=2.5}$ and $|\phi(f(y))-\phi(x)|_{\sigma_{max}=1.0}$ increases rapidly, while the fidelity gap remains nearly unchanged.
This indicates that perceptual quality is more sensitive to the noise level change, with higher noise levels being particularly important in edge and texture areas.
These regions lie in the low-density regions of the real image distribution and are affected by the ill-posed nature of super-resolution.
Applying a low $\sigma_{max}$ in such areas leads to unreliable score estimation, resulting in perceptually poor, over-smoothed outputs.

The experimental results support our hypothesis that different regions of an image respond differently to noise and require distinct handling based on their content.
Flat regions typically closely resemble the corresponding ground truth, requiring only a slight amount of noise; therefore, we consider them as existing at small timesteps ($t\to 0$).
Based on this idea, we propose to reduce noise levels applied in flat regions, and the results shown by the red line in Fig.~\ref{fig: res} preliminarily verify the effectiveness.
In the following subsections, we will detailedly discuss how to determine the noise level for different types of areas in the LR image.

\subsection{Uncertainty Estimation}
\label{sec: uncertainty estimation}

\begin{figure}[t]
\centering
\vspace{1.5mm}
\includegraphics[width=\linewidth]{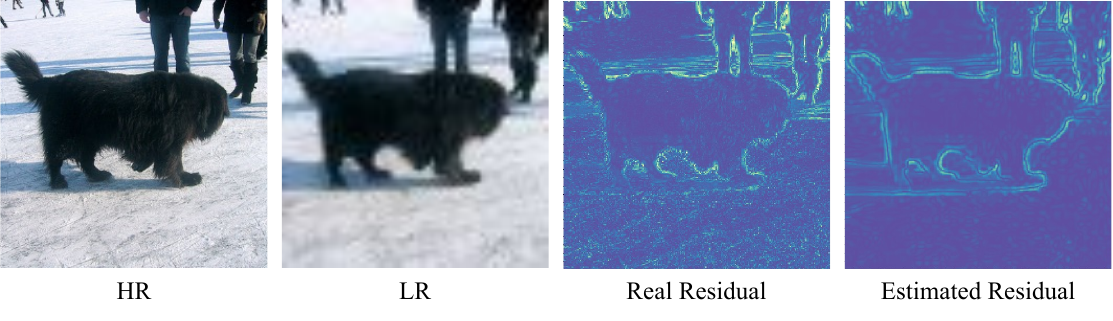}

\vspace{-3mm}
\caption{A visualization of the actual residual $|\bm{x}^i - \bm{y}^i|$ and the estimated residual $|g(\bm{y}^i) - \bm{y}^i|$. The real residual exhibits high values in edges and texture regions, indicating the high uncertainty. The residual estimated by SR network is close to the real one and therefore can serve as a rough estimation of uncertainty.}
\label{fig: vis un}
\vspace{-4mm}
\end{figure}

Due to the ill-posed nature of the image super-resolution task, perfectly reconstructing HR images from degraded LR inputs is non-trivial.
The high-frequency information in edge and texture areas is severely corrupted during the degradation, exhibiting greater variance compared to flat areas and making accurate prediction more challenging.
Several works~\cite{lee2019gram, ning2021uncertainty} explore this variance from an uncertainty-based perspective~\cite{kendall2017uncertainties}.
Given an HR image $\bm{x}^i\in\mathbb{R}^{n}$ and its corresponding LR image $\bm{y}^i\in\mathbb{R}^{n}$, the uncertainty $\bm{\psi}^i$ of the SR estimate $g(\bm{y}^i)$ is related to its residual with $\bm{x}^i$:
\begin{equation}
\label{eq: uncertainty_sr}
    \bm{x}^i = g(\bm{y}^i) + \epsilon\ \bm{\psi}(g(\bm{y}^i)),
\vspace{-0.5mm}
\end{equation}
where $\epsilon$ represents a standard Laplace or Gaussian distribution when $g(\cdot)$ is regularized by $L1$ or $L2$ loss function, respectively.
If the reconstruction error $|\bm{x}^i - g(\bm{y}^i)|$ is larger than $|\bm{x}^j - g(\bm{y}^j)|$, then it is more likely that $g(\bm{y}^i)$ has a higher uncertainty $\bm{\psi}(g(\bm{y}^i))$, compared to the uncertainty $\bm{\psi}(g(\bm{y}^j))$ of $g(\bm{y}^j)$.
Similarly, we can associate the uncertainty of $\bm{y}^i$ with the residual $|\bm{x}^i - \bm{y}^i|$ as:
\vspace{-0.5mm}
\begin{equation}
\label{eq: uncertainty_lr}
    \bm{x}^i = \bm{y}^i + \hat{\epsilon}\ \bm{\psi}(\bm{y}^i),
\vspace{-0.5mm}
\end{equation}
where $\hat{\epsilon}$ represents an unknown distribution that depends on the degradation pattern.
If $g(\cdot)$ is well-trained, we can assume that $g(\bm{y}^i)$ closely approximates $\bm{x}^i$, implying that $|g(\bm{y}^i) - \bm{y}^i|$ is similar to $|\bm{x}^i - \bm{y}^i|$.
As illustrated in Fig.~\ref{fig: vis un}, the residual estimated by $g(\cdot)$ is roughly similar to the real residual.
Therefore, we propose to leverage the residual $|g(\bm{y}^i) - \bm{y}^i|$ as the estimate of the uncertainty of $\bm{y}^i$.
Specifically, we define the uncertainty estimate as:
\vspace{-0.5mm}
\begin{equation}
\label{eq: uncertainty_est}
    \bm{\psi}_{est}(\bm{y}) = \frac{1}{2}|g(\bm{y}) - \bm{y}|.
\vspace{-0.5mm}
\end{equation}
In the next subsection, we will take this uncertainty estimate as the criterion to adjust the noise intensity.



\subsection{Uncertainty-guided Noise Weighting}
\label{sec: uncertainty-based noise}

\begin{figure*}[t]
\centering
\includegraphics[width=.75\textwidth]{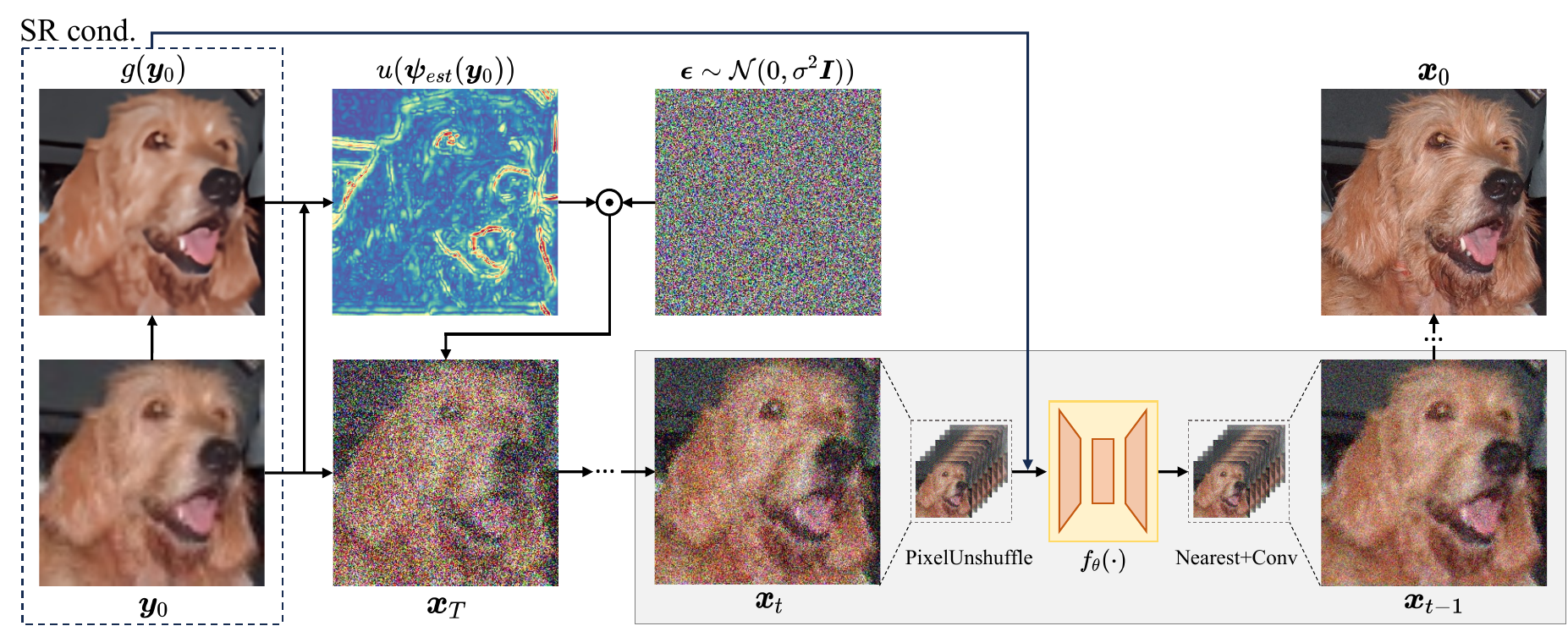}

\vspace{-2mm}
\caption{The overall pipeline of the proposed UPSR model. An auxiliary SR network is first employed to estimate the uncertainty of the input $\bm{y}_0$. Then the weighting coefficient $w_u$ computed based on the uncertainty $\bm{\psi}_{est}(\bm{y}_0)$ are applied to adjust the noise level in different regions. Meanwhile, both the SR estimate $g(\bm{y}_0)$ and LR input $\bm{y}_0$ are concatenated as the conditional information for the denoiser $f_\theta(\cdot)$.}
\label{fig: UWSR pipeline}
\vspace{-3mm}
\end{figure*}

Based on the discussion in previous sections, we propose to replace the commonly used isotropic Gaussian noise with anisotropic noise that adapts to the image content.
Inspired by \cite{ning2021uncertainty}, which deals with areas with higher uncertainty by assigning larger weights to the loss function to impose stronger constraints, we suggest adjusting the noise intensity based on the uncertainty estimate across different regions.
Specifically, after obtaining the uncertainty estimate as outlined in Sec.~\ref{sec: uncertainty estimation}, we compute the weighting coefficient $w_u$ and apply it to modulate noise levels for different areas in the diffusion process.
We refer to this strategy as Uncertainty-guided Noise Weighting (UNW).
For $\bm{y}$ with lower uncertainty (e.g. in flat areas), it is more likely that $\bm{y}$ resides in high-density regions of the HR data distribution.
In such cases, a lower noise level is sufficient to achieve perceptual quality comparable to that of a higher noise level, while also preserving more LR details, as discussed in Sec.~\ref{sec: influence}.
Conversely, if $\bm{y}$ exhibits high uncertainty (e.g., in edge or texture areas), a higher noise level is requisite to account for the significant distribution disparity and to provide greater chances of reconstructing photo-realistic details.
Therefore, we model the noise weighting coefficient $w_u$ as a monotonically increasing function with respect to the uncertainty: 
\vspace{-0.5mm}
\begin{equation}
    w_u(\bm{y}) := u(\bm{\psi}_{est}(\bm{y})),
\vspace{-0.5mm}
\end{equation}
where $w_u(\bm{y})\in\mathbb{R}^{n\times n}$ represents the diagonal weight matrix used to lower the noise level in low-uncertainty areas while maintaining a higher noise level in high-uncertainty regions, creating a more specialized and adaptive diffusion pipeline for SR.
Details of the implementation of $u(\cdot)$ are presented in the supplementary material.

Built upon the UNW technique, we introduce a new pipeline for diffusion-based SR, termed Uncertainty-guided Perturbation for SR (UPSR), as illustrated in Fig.~\ref{fig: UWSR pipeline}.
Given an LR image $\bm{y}_0$, we first obtain its SR estimate $g(\bm{y}_0)$ through an auxiliary SR network $g(\cdot)$.
Then we estimate the uncertainty of $\bm{y}_0$ as $\bm{\psi}_{est}(\bm{y}_0) = \frac{1}{2}|g(\bm{y}_0) - \bm{y}_0|$ and obtain the weighting coefficient as $w_u(\bm{y}_0) = u(\bm{\psi}_{est}(\bm{y}_0))$.
Based on the weighting coefficient, we rewrite the forward transition distribution $q(\bm{x}_t \mid \bm{x}_{t-1}, \bm{x}_{0}, \bm{y}_0)$ in Eq.~\ref{eq: resshift forward transition} as:
\vspace{-0.5mm}
\begin{equation}
\label{eq: rewritten forward transition}
    \mathcal{N} \left (\bm{x}_t \mid \bm{x}_{t-1} + \alpha_t (\bm{y}_{0} - \bm{x}_0), \kappa^2 w_u(\bm{y}_0)^2 \alpha_t \bm{I} \right ), 
\vspace{-0.5mm}
\end{equation}
and derive the corresponding backward transition distribution $q(\bm{x}_{t-1} \mid \bm{x}_{t}, \bm{x}_{0}, \bm{y}_0)$ as:
\vspace{-0.5mm}
\begin{equation}
\label{eq: rewritten backward transition}
    \mathcal{N} \left ( \bm{x}_{t-1} \mid \frac{\eta_{t-1}}{\eta_t} \bm{x}_{t} + \frac{\alpha_t}{\eta_t} \bm{x}_0, \kappa^2 w_u(\bm{y}_0)^2 \frac{\eta_{t-1}}{\eta_t} \alpha_t \bm{I} \right ),
\vspace{-0.5mm}
\end{equation}
where the difference is that we leverage the weighting coefficient $w_u(\bm{y}_0)$ to control the noise level.
Since $g(\bm{y}_0)$ is a better estimate of $\bm{x}_0$ compared to $\bm{y}_0$, we combine $g(\bm{y}_0)$ with $\bm{y}_0$ to provide more accurate conditional information for the denoiser.
Details of the derivations of Eq.~\ref{eq: rewritten forward transition}, \ref{eq: rewritten backward transition} and the training pipeline are presented in the supplementary materials.
Following the previous diffusion-based methods~\cite{ho2020denoising, yue2024resshift}, our training objective for the denoiser to predict the target $\bm{x}_0$ combines both pixel distance $||\cdot||_2^2$ and LPIPS criterion $L_{per}$:
\vspace{-1mm}
\begin{equation}
\begin{split}
    \mathcal{L} (\theta) = \sum_t [ & || f_\theta(\bm{x}_t, \bm{y}_0, g(\bm{y}_0), t) - \bm{x}_0 ||_2^2 \\
    &+ \lambda L_{per} (f_\theta(\bm{x}_t, \bm{y}_0, g(\bm{y}_0), t), \bm{x}_0) ], 
\end{split}
\vspace{-1mm}
\end{equation}
where $\lambda$ is a hyperparameter to control the trade-off between fidelity and perceptual quality. Optimizing the denoiser $f_\theta(\cdot)$ through the mixed objective function facilitates fewer diffusion steps while achieving better photo-realistic results in various benchmark real-world datasets.

\begin{table*}
\small
\setlength{\tabcolsep}{5pt}

\caption{Ablation study on effects of the proposed components, including the SR condition and uncertainty-guided noise weighting (UNW). The best results are highlighted in \textbf{bold}.}
\label{tab: ablation}

\vspace{-3mm}
  \begin{center}
  \begin{tabular}{cc|ccccc|cccc}
    \toprule
    \multirow{2}{*}{UNW} & \multirow{2}{*}{SR cond.} & \multicolumn{5}{c|}{RealSR} & \multicolumn{4}{c}{RealSet} \\
    & & PSNR$\uparrow$ & CLIPIQA$\uparrow$ & MUSIQ$\uparrow$ & MANIQA$\uparrow$ & NIQE$\downarrow$ & CLIPIQA$\uparrow$ & MUSIQ$\uparrow$ & MANIQA$\uparrow$ & NIQE$\downarrow$ \\

    \midrule
            &         & 26.18 & 0.5447 & 62.951 & 0.3596 & 4.49 & 0.6141 & \textbf{64.360} & 0.3718 & 4.42 \\
    $\surd$ &         & 26.12 & 0.5760 & 64.512 & 0.3717 & 4.18 & 0.6340 & 64.280 & 0.3836 & \textbf{4.22} \\
    $\surd$ & $\surd$ & \textbf{26.44} & \textbf{0.6010} & \textbf{64.541} & \textbf{0.3818} & \textbf{4.02} & \textbf{0.6389} & 63.498 & \textbf{0.3931} & 4.24 \\

    \bottomrule
  \end{tabular}
  \end{center}
\vspace{-5mm}
\end{table*}

\subsection{Network Architecture Modification}
\label{sec: architecture}
Besides altering the noise injection scheme and sampling procedure, we also make several modifications to the network architecture.

We rethink the necessity of leveraging latent space for image super-resolution with only a few diffusion steps.
\cite{rombach2022high} and \cite{yue2024resshift} utilize VQGAN~\cite{esser2021taming} to transfer the diffusion process from pixel space to latent space, for the sake of reducing spatial dimensionality to improve efficiency.
However, as the number of sampling steps decreases, the performance-cost ratio of VQGAN gradually decreases due to its own huge computing resource consumption.
Moreover, the application of certain perceptual criteria (e.g., LPIPS~\cite{zhang2018unreasonable}) requires pixel-level computation, leading to additional decoding costs during the training process.
To mitigate its impact, we propose to replace the VQGAN encoder and decoder with PixelUnshuffle operation~\cite{shi2016real} and a simple nearest neighbor upsampling module, respectively. This approach enables compressing and expanding the spatial dimension with minimal computational cost.
Therefore, we can perform the diffusion process directly in pixel space while maintaining a complexity comparable to that in the latent space. 
%
In this vein, we significantly reduce the total model size and training overhead without sacrificing performance.

\section{Experiments}
\label{experiments}

\subsection{Experimental Setups}

For data preparation, we apply randomly cropped $256\times 256$ patches from ImageNet~\cite{deng2009imagenet} as HR training data following \cite{rombach2022high} and \cite{yue2024resshift}.
Then the degradation pipeline of RealESRGAN~\cite{wang2021real} are adopted to generate degraded $64\times 64$ LR inputs.
For network architecture, we employ the commonly-used U-Net denoiser $f_\theta(\cdot)$ with few adjustments to fit our refined diffusion pipeline.
For $g(\cdot)$, we employ the lightweight version of \cite{zhang2024transcending} and pretrain it following the real SR pipeline to produce SR predictions and uncertainty estimates.
During the diffusion model training process, we freeze the pretrained $g(\cdot)$ and optimize $f_\theta(\cdot)$ for 200k iterations with a batchsize of 32.
To quantitatively validate the efficacy of the proposed method, we utilize three full-reference metrics: PSNR, SSIM and LPIPS~\cite{zhang2018unreasonable}; and four non-reference metrics CLIPIQA~\cite{wang2023exploring}, MUSIQ~\cite{ke2021musiq}, MANIQA~\cite{yang2022maniqa}, and NIQE~\cite{mittal2012making} to evaluate the performance.
Among these metrics, PSNR and SSIM reflect the fidelity of the generated super-resolution results, while other metrics assess the perceptual quality of the outputs.

\subsection{Analysis}

In this subsection, we conduct ablation studies on the effectiveness of several key components and analyze the performance-cost trade-off of the proposed method.

\paragraph{Effectiveness of UNW and SR conditioning.}
In order to show the efficacy of the proposed uncertainty-guided noise weighting scheme and SR conditioning, we use the noise schedule and sampling process of ResShift~\cite{yue2024resshift} as the baseline.
We first apply the uncertainty-guided noise weighting scheme to adjust the noise intensity across different regions based on the uncertainty estimate.
The weighting scheme yields better outcomes in terms of perceptual quality, with enhancements of 0.0355 in CLIPIQA and 1.512 in MUSIQ.
We attribute this improvement to the reduced noise intensities applied to flat areas, which allow these clearer regions to provide additional information that aids in more accurately recovering the remaining degraded regions.
This validates our idea of reducing noise in flat regions does not degrade perceptual quality, but rather enhances it.
Next, we incorporate the SR prediction $g(\bm{y}_0)$ and LR input $\bm{y}_0$ as the conditional information entered into the denoiser $f_\theta(\cdot)$. 
The additional SR prediction offers more precise supplementary information compared to relying solely on the degraded LR input, resulting in a significant improvement in both fidelity and perceptual quality.
Specifically, this more robust pipeline further improves PSNR by 0.34 dB and CLIPIQA by 0.0247 in the RealSR dataset.
These ablation studies verify the significance of leveraging LR information during diffusion-based SR process.


\begin{table}
\footnotesize
\setlength{\tabcolsep}{5pt}

\vspace{1mm}
\caption{Model size and computational efficiency comparisons between the proposed UPSR and other diffusion-based methods. \gray{Gray} numbers denote the parameter counts for auxiliary network, i.e., VQGAN in ResShift and $g(\cdot)$ in our work. We test the runtimes on $32\times 3\times 64\times 64$ LR input using single RTX4090 GPU and present several results evaluated on the RealSR dataset.
}
\label{tab: model size comparison}
\vspace{-3mm}

\begin{center}
\begin{tabular}{l|cc|cc}
\toprule
\textbf{Model} & Params (M) & Runtimes (s) & MUSIQ & MANIQA \\  
\midrule
LDM-15      & 113.60+\gray{55.32} & 1.59 & 48.698 & 0.2655 \\
ResShift-15 & 118.59+\gray{55.32} & 1.98 & 57.769 & 0.3691 \\  
ResShift-4  & 118.59+\gray{55.32} & 1.00 & 55.189 & 0.3337 \\  
UPSR-5      & 119.42+\gray{2.50}  & 1.12 & 64.541 & 0.3818 \\

\bottomrule
\end{tabular}
\end{center}
  
\end{table}

\begin{table}[]
\centering
\footnotesize
\setlength{\tabcolsep}{5pt}
\vspace{-4mm}
\caption{Training overhead comparison between ResShift and UPSR evaluated under a batchsize of 16 per GPU.}
\label{tab: cost}
\vspace{-1mm}
\begin{tabular}{l|cc}
\toprule
Model & Training Speed & Memory Footprint \\
\midrule
ResShift   & 1.20 s/iter & 24.1 G \\
UPSR       & 0.45 s/iter & 14.9 G \\
\bottomrule
\end{tabular}
\vspace{-3mm}
\end{table}
\begin{table*}
\setlength{\tabcolsep}{3pt}
\footnotesize
\caption{Quantitative results of different methods on one synthetic dataset ImageNet-Test and two real-world datasets RealSR and RealSet. The best and second best results are marked in \red{red} and \blue{blue}. 
All results of the previous methods are evaluated using the released inference codes and pretrained weights.
}
\label{tab: sota results}
\vspace{-5mm}
  \begin{center}
  \begin{tabular}{c|c|cccccc|cccc}
    \toprule
    \multirow{2}{*}{Datasets} & \multirow{2}{*}{Metrics} & \multicolumn{6}{c|}{GAN-based Methods} & \multicolumn{4}{c}{Diffusion-based Methods} \\
    & & ESRGAN & RealSR-JPEG & BSRGAN & RealESRGAN & $\ \ $SwinIR$\ \ $ & $\ \ $DASR$\ \ $ & LDM-15 & ResShift-15 & ResShift-4 & UPSR-5  \\
    \midrule
    
    \multirow{7}{*}{\makecell{\textit{ImageNet}\\\textit{-Test}}}
    & PSNR$\uparrow$    & 20.67 & 23.11 & \blue{24.42} & 24.04 & 23.99 & \red{24.75} & 24.85 & \blue{24.94} & \red{25.02} & 23.77\\
    & SSIM$\uparrow$    & 0.4485 & 0.5912 & 0.6585 & 0.6649 & \blue{0.6666} & \red{0.6749} & 0.6682 & \blue{0.6738} & \red{0.6830} & 0.6296 \\
    & LPIPS$\downarrow$ & 0.4851 & 0.3263 & 0.2585 & 0.2539 & \red{0.2376} & \blue{0.2498} & 0.2685 & \blue{0.2371} & \red{0.2075} & 0.2456 \\
    & CLIPIQA$\uparrow$ & 0.4512 & 0.5366 & \red{0.5810} & 0.5241 & \blue{0.5639} & 0.5362 & 0.5095 & 0.5860 & \blue{0.6003} & \red{0.6328} \\
    & MUSIQ$\uparrow$   & 43.615 & 46.981 & \red{54.696} & 52.609 & \blue{53.789} & 48.337 & 46.639 & \blue{53.182} & 52.019 & \red{59.227} \\
    & MANIQA$\uparrow$  & 0.3212 & 0.3065 & \blue{0.3865} & 0.3689 & \red{0.3882} & 0.3292 & 0.3305 & \blue{0.4191} & 0.3885 & \red{0.4591} \\
    & NIQE$\downarrow$  & 8.33 & 5.96 & 6.08 & 6.07 & \blue{5.89} & \red{5.86} & 7.21 & \blue{6.88} & 7.34 & \red{5.24} \\
    \midrule
    \multirow{7}{*}{\textit{RealSR}}
    & PSNR$\uparrow$    & \red{27.57} & \blue{27.34} & 26.51 & 25.83 & 26.43 & 27.19 & \red{27.18} & \blue{26.80} & 25.77 & 26.44 \\
    & SSIM$\uparrow$    & 0.7742 & 0.7605 & 0.7746 & 0.7726 & \red{0.7861} & \red{0.7861} & \red{0.7853} & \blue{0.7674} & 0.7439 & 0.7589 \\
    & LPIPS$\downarrow$ & 0.4152 & 0.3962 & \blue{0.2685} & 0.2739 & \red{0.2515} & 0.3113 & \blue{0.3021} & 0.3411 & 0.3491 & \red{0.2871} \\
    & CLIPIQA$\uparrow$ & 0.2362 & 0.3613 & \red{0.5439} & \blue{0.4923} & 0.4655 & 0.3628 & 0.3748 & \blue{0.5709} & 0.5646 & \red{0.6010} \\
    & MUSIQ$\uparrow$   & 29.037 & 36.069 & \red{63.587} & \blue{59.849} & 59.635 & 45.818 & 48.698 & \blue{57.769} & 55.189 & \red{64.541} \\
    & MANIQA$\uparrow$  & 0.2071 & 0.1783 & \red{0.3702} & \blue{0.3694} & 0.3436 & 0.2663 & 0.2655 & \blue{0.3691} & 0.3337 & \red{0.3828} \\
    & NIQE$\downarrow$  & 7.73 & 6.95 & \red{4.65} & \blue{4.68} & \blue{4.68} & 5.98 & 6.22 & \blue{5.96} & 6.93 & \red{4.02} \\
    \midrule
    \multirow{4}{*}{\textit{RealSet}}
    & CLIPIQA$\uparrow$ & 0.3739 & 0.5282 & \red{0.6160} & \blue{0.6081} & 0.5778 & 0.4966 & 0.4313 & \blue{0.6309} & 0.6188 & \red{0.6392} \\
    & MUSIQ$\uparrow$   & 42.366 & 50.539 & \red{65.583} & \blue{64.125} & 63.817 & 55.708 & 48.602 & \blue{59.319} & 58.516 & \red{63.519} \\
    & MANIQA$\uparrow$  & 0.3100 & 0.2927 & \blue{0.3888} & \red{0.3949} & 0.3818 & 0.3134 & 0.2693 & \blue{0.3916} & 0.3526 & \red{0.3931} \\
    & NIQE$\downarrow$  & 4.93 & 4.81 & 4.58 & \red{4.38} & \blue{4.40} & 4.72 & 6.47 & \blue{5.96} & 6.46 & \red{4.23} \\

    \bottomrule
  \end{tabular}
  \end{center}
\vspace{-6mm}
  
\end{table*}


\paragraph{Model Size and Training Overhead Comparison.}
In section~\ref{sec: architecture}, we discussed replacing VQGAN in diffusion-based SR model with simpler downsampling and upsampling modules to improve efficiency.
To validate the efficacy of the modified architecture, we first evaluate the model size and computational cost of UPSR compared to state-of-the-art diffusion-based methods, as shown in Tab.~\ref{tab: model size comparison}.
With the modified architecture and pipeline, UPSR-5 reduces the overall model size by 30\% and achieves better perceptual quality with an inference speed comparable to that of ResShift-4.
In contrast, ResShift-4 achieves the acceleration at the expense of performance.
We then assess the training overhead of different methods, as shown in Tab.~\ref{tab: cost}.
Removing VQGAN results in substantial reductions in training overhead, increasing training speed by 167\% and saving GPU memory footprint by 38\%.
These results validate that UPSR strikes a better trade-off between performance and efficiency.





\subsection{Comparisons with State-of-the-Art Methods}

We select ImageNet-Test~\cite{deng2009imagenet, yue2024resshift} that contains 3,000 images as major dataset for synthetic image super-resolution evaluation.
Furthermore, two commonly used real-world datasets, RealSR~\cite{cai2019toward} and RealSet65~\cite{yue2024resshift}, are adopted to evaluate the generalizability in real-world scenarios.
We make comparisons with several GAN-based methods ESRGAN~\cite{wang2018esrgan}, RealSR-JPEG~\cite{ji2020real}, BSRGAN~\cite{zhang2021designing}, RealESRGAN~\cite{wang2021real}, SwinIR~\cite{liang2021swinir}, DASR~\cite{liang2022efficient}, and two diffusion-based methods LDM~\cite{rombach2022high} and ResShift~\cite{yue2024resshift}.

The quantitative results on three benchmark datasets are shown in Tab.~\ref{tab: sota results}.
We present UPSR with five sampling steps in order to trade off the super-resolution performance and computational consumption.
The proposed UPSR achieves better CLIPIQA, MUSIQ, MANIQA and NIQE compared to ResShift, indicating the significant improvement in perceptual quality.
Specifically, while reducing 30\% of the overall model size, UPSR-5 still outperforms ResShift-4 by 7.21 and 2.10 in terms of MUSIQ and NIQE in ImageNet-Test dataset.
For real-world datasets, our method also yields impressive results. 
UPSR-5 consistently outperforms ResShift-4 across both the RealSR and RealSet datasets, demonstrating the efficacy of UPSR in addressing unknown degradation. 
Furthermore, UPSR-5 achieves the best NIQE metric, while recent diffusion-based methods show poorer NIQE performance compared to most GAN-based methods.

\subsection{Visual Examples}

\begin{figure}[tbp]
    \centering
    \vspace{1mm}
    
    \includegraphics[width=.95\linewidth]{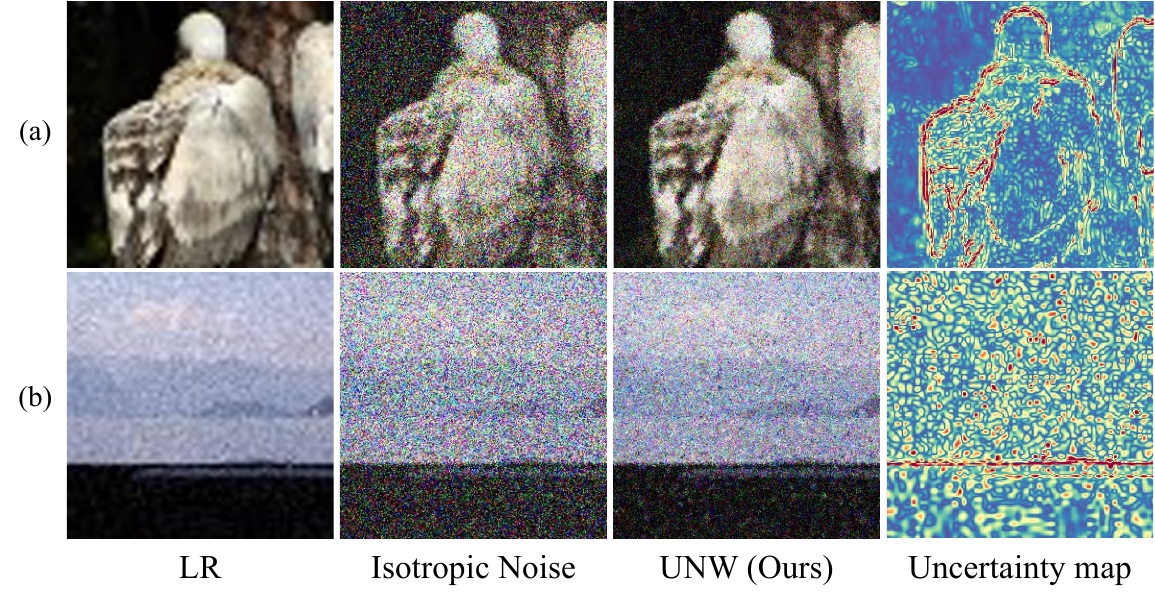}
    
    \vspace{-2.5mm}
    \caption{Visual examples of the proposed UNW strategy. Based on the uncertainty estimate (illustrated as the heatmap), the noise level in most flat areas is reduced to preserve more details for better SR results. Meanwhile, noise in edge areas (e.g., in image (a)) and severely degraded parts (e.g., in image (b)) are maintained relatively heavy to ensure reliable score estimation to produce visually pleasing results. }  
    \label{fig: vis of noise}
    \vspace{-5mm}
\end{figure}

\begin{figure*}[tbp]
    \centering
    \includegraphics[width=.8\linewidth]{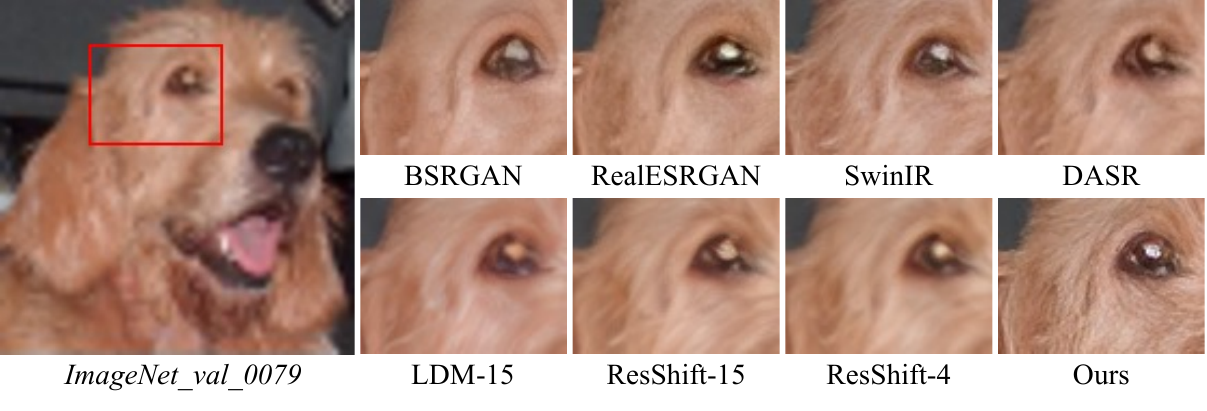}

    \vspace{-2mm}
    
    \includegraphics[width=.8\linewidth]{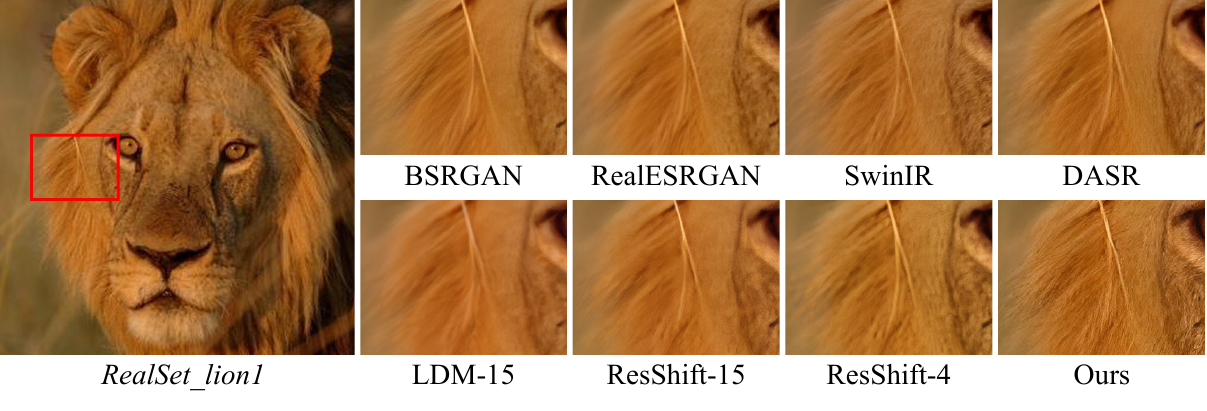}

    \vspace{-2mm}

    \includegraphics[width=.8\linewidth]{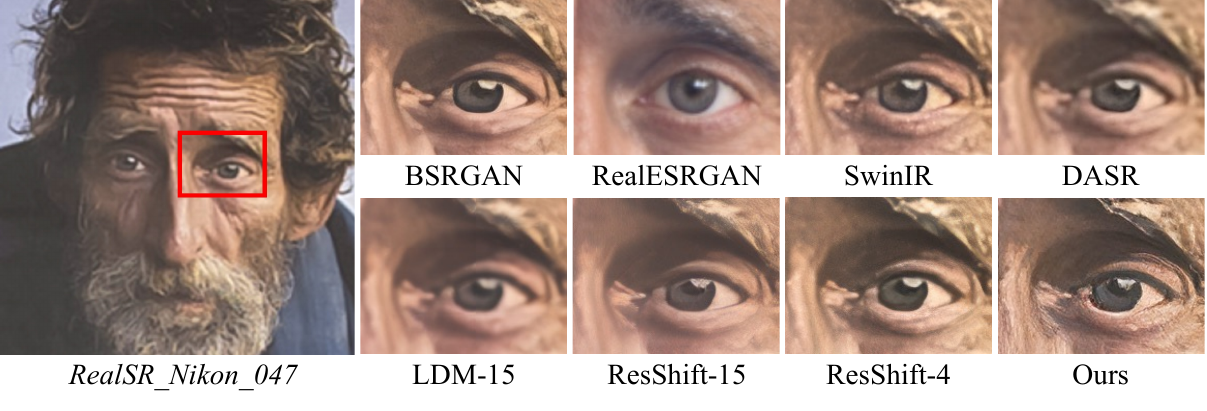}
    
    \vspace{-2.5mm}
    \caption{Qualitative comparison between different methods on real-world datasets. \textit{Please zoom in for more details.}}
    \label{fig: vis comp}
    \vspace{-3mm}
\end{figure*}

In this section, we present several visual examples to illustrate the proposed UNW scheme and the comparisons between different methods.

\paragraph{Visualization of Weighted Noise.}
\label{sec: vis}
In Fig.~\ref{fig: vis of noise} we provide visual examples of isotropic and anisotropic noises applied in different images.
The proposed UNW strategy enables adaptive perturbation across different areas, leading to the anisotropic diffusion process.
The reduced noise level in flat areas, particularly in clear backgrounds, help retain more details in the initial state without sacrificing generative capability.
Additionally, these retained details provide more information to reconstruct missing details in surrounding regions, thereby enhancing overall performance.
Conversely, in edge regions, the noise level is maintained at the predefined $\sigma_{max}$ due to their high uncertainty, ensuring sufficient perturbation to prevent the model from generating over-smoothed results.

\vspace{-1mm}
\paragraph{Qualitative Comparison.}
\label{sec: comp}

In Fig.~\ref{fig: vis comp}, we provide visual results produced by different methods on real-world datasets. 
These results validate that the proposed UPSR method is capable of producing visually better outputs.
Specifially, UPSR successfully recover finer textures and sharper edges, while other methods yields blurry results. 
More visual examples are presented in the supplementary material.

\section{Conclusion}
In this work, we propose specialized prior distribution and sampling pipeline, along with modifications on the network architecture to develop the Uncertainty-guided Perturbation scheme for Super-Resolution to tackle real-world SR task.
Specifically, we propose region-specific processing based on the LR content from a uncertainty-based perspective.
In flat areas (low-uncertainty), reducing the noise level could preserve more details in the initial state of the diffusion process and therefore improve performance,
whereas edge and texture regions (high-uncertainty) demand stronger noise to bridge their significant disparity with the target distribution.
To facilitate region-specific control, we leverage the uncertainty of LR image as the criterion to adjust noise level in different areas.
We employ an auxiliary SR network and consider the residual of SR prediction as the approximate uncertainty estimate of LR image due to their similarity.
We further incorporate the SR prediction as conditional information and make several architectural modifications to provide a more robust pipeline to further enhance performance. 
Extensive experiments on both synthetic and real-world datasets validate the efficacy of the proposed UPSR model, even with lower model size and training overhead. 

\paragraph{Acknowledgement.} This work was supported by National Natural Science Foundation of China (No. 62476051) and Sichuan Natural Science Foundation (No. 2024NSFTD0041).



{
    \small
    \bibliographystyle{ieeenat_fullname}
    \bibliography{main}
}
\clearpage
\setcounter{section}{0}
\renewcommand\thesection{\Alph{section}}
\maketitlesupplementary

In this supplementary material, we present more implementation details of the proposed UPSR method and additional visual results.
Firstly, we introduce the details of the sampling process of UPSR in Sec.~\ref{sec:sampling}.
Secondly, we present the details of weighting coefficient $u(\cdot)$ in Sec.~\ref{sec:weighting coefficient}.
Then, comparisons to several existing methods are made in Sec.~\ref{sec:comparison1} and Sec.~\ref{sec:comparison2}.
Lastly, additional visual examples are shown in Sec.~\ref{sec:more vis}.

\section{Details of the Sampling Process}
\label{sec:sampling}
\paragraph{Derivation of Eq.~\ref{eq: rewritten forward transition},~\ref{eq: rewritten backward transition}:}
As discussed in Sec.~\ref{sec: uncertainty-based noise}, we apply region-specific weighting coefficient to the noise level based on the sampling process of ResShift~\citep{yue2024resshift} and replace $\bm{x}_t = \bm{x}_{t-1} + \alpha_t (\bm{y}_{0} - \bm{x}_0) + \kappa \sqrt{\alpha_t} \bm{\xi}_t$ with $\bm{x}_t = \bm{x}_{t-1} + \alpha_t (\bm{y}_{0} - \bm{x}_0) + \kappa \underline{w_u(\bm{y}_0)} \sqrt{\alpha_t} \bm{\xi}_t$, where $\bm{\xi}_t \sim \mathcal{N}(\bm{0}, \bm{I})$ is from independently distributed Gaussian distribution and $\alpha$ is a scaling factor. Therefore, we rewrite the forward transition distribution as:
\begin{equation}
\label{eq: forward transition}
    q(\bm{x}_t \mid \bm{x}_{t-1}, \bm{x}_{0}, \bm{y}_0) = \mathcal{N} \left (\bm{x}_t \mid \bm{x}_{t-1} + \alpha_t (\bm{y}_{0} - \bm{x}_0), \kappa^2 w_u(\bm{y}_0)^2 \alpha_t \bm{I} \right ).
\end{equation}
Areas with higher uncertainty, e.g. edge and texture areas, are assigned larger weighting coefficients $w_u(\bm{y}_0)$ and therefore greater noise $\kappa w_u(\bm{y}_0) \sqrt{\alpha_t} \bm{\xi}_t$.
Meanwhile, $\bm{x}_t$ can be reparameterized as:
\begin{equation}
\begin{split}
    \bm{x}_t &= \bm{x}_0 + \sum_{i=1}^t (\bm{x}_i - \bm{x}_{i-1}) = \bm{x}_0 + \sum_{i=1}^t (\alpha_t (\bm{y}_{0} - \bm{x}_0) + \kappa w_u(\bm{y}_0) \sqrt{\alpha_t} \bm{\xi}_t) \\
    &= \bm{x}_0 + \eta_t (\bm{y}_{0} - \bm{x}_0) + \kappa w_u(\bm{y}_0) \sqrt{\eta_t} \bm{\xi}_t,
\end{split}
\end{equation}
leading to the marginal distribution at time $t$ as:
\begin{equation}
\label{eq: forward marginal}
    q(\bm{x}_t \mid \bm{x}_{0}, \bm{y}_0) = \mathcal{N} \left (\bm{x}_t \mid \bm{x}_{0} + \eta_t (\bm{y}_{0} - \bm{x}_0), \kappa^2 w_u(\bm{y}_0)^2 \eta_t \bm{I} \right ).
\end{equation}
According to Bayes's theorem, the reverse transition distribution can be written as 
\begin{equation}
\label{eq: bayesian}
\begin{split}    
    q(\bm{x}_{t - 1} \mid \bm{x}_{t}, \bm{x}_{0}, \bm{y}_0) &= \frac{ q(\bm{x}_{t} \mid \bm{x}_{t - 1}, \bm{x}_{0}, \bm{y}_0) q(\bm{x}_{t - 1} \mid \bm{x}_{0}, \bm{y}_0) }{q(\bm{x}_{t} \mid \bm{x}_{0}, \bm{y}_0)} \\
    &\propto q(\bm{x}_{t} \mid \bm{x}_{t - 1}, \bm{x}_{0}, \bm{y}_0) q(\bm{x}_{t - 1} \mid \bm{x}_{0}, \bm{y}_0).
\end{split}
\end{equation}
Incorporating Eq.~\ref{eq: forward transition}, Eq.~\ref{eq: forward marginal}, and Eq.~\ref{eq: bayesian}, we consider the distribution at each pixel $i$ as:
\begin{equation}
\label{eq: X}
\begin{split}    
    q(x^{i}_{t-1}|x^{i}_t,x^{i}_0,y^{i}_0)
    &\propto q(x^{i}_{t} \mid x^{i}_{t - 1}, x^{i}_{0}, y^{i}_0) q(x^{i}_{t - 1} \mid x^{i}_{0}, y^{i}_0) \\
    &\propto \operatorname{exp}\left( -\frac{(x^{i}_t - \mu^{i}_1)^2}{2\kappa^2 w_u(y^{i}_0)^2 \alpha_t} \right) \operatorname{exp}\left( -\frac{(x^{i}_{t-1} - \mu^{i}_2)^2}{2\kappa^2 w_u(y^{i}_0)^2 \eta_{t-1}} \right) \\
    &= \operatorname{exp}\left( -\frac{\eta_{t-1}(x^{i}_{t-1} - \mu^{i}_3)^2 + \alpha_t (x^{i}_{t-1} - \mu^{i}_2)^2}{2\kappa^2 w_u(y^{i}_0)^2 \alpha_t \eta_{t-1}} \right) \\
    &= \operatorname{exp}\left( -\frac{\eta_{t}(x^{i}_{t-1})^2 - x^{i}_{t-1}(\eta_{t-1}\mu^{i}_3 + \alpha_t\mu^{i}_2) - (\eta_{t-1}\mu^{i}_3 + \alpha_t\mu^{i}_2)x^{i}_{t-1}}{2\kappa^2 w_u(y^{i}_0)^2 \alpha_t \eta_{t-1}} \right),
\end{split}
\end{equation}
where $\mu^{i}_1=x^{i}_{t-1} + \alpha_t (y^{i}_{0} - x^{i}_0)$, $\mu^{i}_2=x^{i}_{0} + \eta_{t-1} (y^{i}_{0} - x^{i}_0)$, and $\mu^{i}_3 = (x^{i}_t-\mu^{i}_1) + x^{i}_{t-1} = x^{i}_{t} - \alpha_t (y^{i}_{0} - x^{i}_0)$. 
Next, we further simplify Eq.~\ref{eq: X} to:
\begin{equation}
\label{eq: X2}
\begin{split}    
    q(x^{i}_{t-1}|x^{i}_t,x^{i}_0,y^{i}_0)
    &\propto \operatorname{exp}\left( -\frac{(x^{i}_{t-1} - \frac{1}{\eta_t}(\eta_{t-1}\mu^{i}_3 + \alpha_t\mu^{i}_2))^2}{2\kappa^2 w_u(y^{i}_0)^2 \alpha_t \eta_{t-1} / \eta_t} + \operatorname{constant} \right) \\
    &= \operatorname{exp}\left( -\frac{(x^{i}_{t-1} - (\frac{\eta_{t-1}}{\eta_t}x^{i}_t+ \frac{\alpha_t}{\eta_t}x^{i}_0))^2}{2\kappa^2 w_u(y^{i}_0)^2 \alpha_t \eta_{t-1} / \eta_t} + \operatorname{constant} \right).
\end{split}
\end{equation}
Therefore, we present the reverse transition distribution in Eq.~\ref{eq: rewritten backward transition} based on Eq.~\ref{eq: X2}:
\begin{equation}
\begin{split}    
    q(\bm{x}_{t-1} \mid \bm{x}_{t}, \bm{x}_{0}, \bm{y}_0) 
    &= \prod_{i} q(x^{i}_{t-1}|x^{i}_t,x^{i}_0,y^{i}_0) \\
    &= \mathcal{N} \left ( \bm{x}_{t-1} \mid \frac{\eta_{t-1}}{\eta_t} \bm{x}_{t} + \frac{\alpha_t}{\eta_t} \bm{x}_0, \kappa^2 w_u(\bm{y}_0)^2 \frac{\eta_{t-1}}{\eta_t} \alpha_t \bm{I} \right ).
\end{split}
\end{equation}

\begin{algorithm}
    \caption{Training procedure of UPSR.}
    \label{alg: training}
    \begin{algorithmic}[1]
    \setstretch{1.10}
        \REQUIRE Diffusion model $f_\theta(\cdot)$, pre-trained SR network $g(\cdot)$
        \REQUIRE Paired training dataset $(X, Y)$
        \WHILE{not converged}
            \STATE sample $\bm{x}_0, \bm{y}_0 \sim (X, Y)$
            \STATE sample $t \sim U(1, T)$
            \STATE compute $g(\bm{y}_0)$
            \STATE $\bm{\psi}_{est}(\bm{y}_0) = \frac{1}{2}|g(\bm{y}_0) - \bm{y}_0|$
            \STATE $w_u(\bm{y}_0) = u(\bm{\psi}_{est}(\bm{y}_0))$
            \STATE sample $\epsilon \sim \mathcal{N}(\bm{0}, \kappa^2 \eta_t w_u(\bm{y}_0)^2 \bm{I})$
            \STATE $\bm{x}_t = \bm{x}_0 + \eta_t (\bm{y}_0 - \bm{x}_0) + \epsilon$
            \STATE $\mathcal{L} (\theta) = \sum_t \left[|| f_\theta(\bm{x}_t, \bm{y}_0, g(\bm{y}_0), t) - \bm{x}_0 ||_2^2 + \lambda L_{per} (f_\theta(\bm{x}_t, \bm{y}_0, g(\bm{y}_0), t), \bm{x}_0) \right]$
            \STATE Take a gradient descent step on $\nabla_\theta \mathcal{L} (\theta)$
        \ENDWHILE
        \RETURN Converged diffusion model $f_\theta(\cdot)$.
    \end{algorithmic}
\end{algorithm}

\paragraph{Details of the training procedure.} We present the detailed training pipeline of the proposed UPSR method in Alg.~\ref{alg: training}.

\section{Implementation of the Weighting Coefficient for Uncertainty-guided Perturbation}
\label{sec:weighting coefficient}

As a supplement to Sec.~\ref{sec: uncertainty-based noise}, we model the relationship between the weighting coefficient of region-specific perturbation and the uncertainty estimate as a monotonically increasing function $u'(\cdot)$ followed by a diagonalization process, i.e., $w_u(\bm{y}_0) = u(\bm{\psi}_{est}(\bm{y}_0)) = \operatorname{diag}(u'(\bm{\psi}_{est}(\bm{y}_0)))$. In this section, we will detailedly introduce the implementation of this weighting coefficient function $u'(\cdot)$ in scalar form.

\begin{figure}[t]
    \centering
    \includegraphics[width=0.65\linewidth]{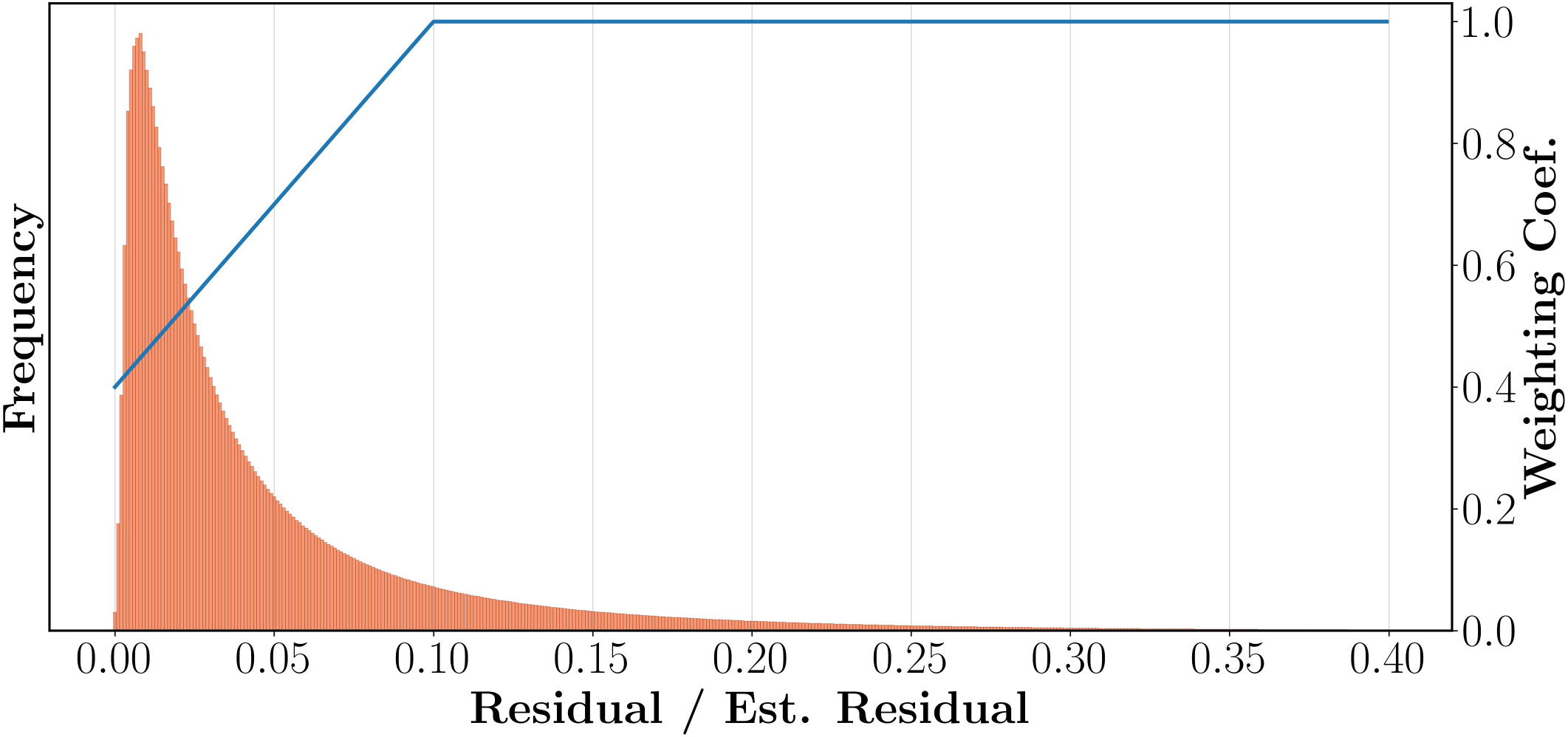}
    \caption{A combined visualization of the distribution of the \textbf{residual} $|y-x|$ (left), and the weighting coefficient $u(\bm{\psi}_{est}(y))$ with respect to the \textbf{estimated residual} $|y-g(y)|$. In the region where the estimated residual is within [0, 0.1] (involving more than 80\% of the data), the value of weight coefficient function increases linearly with the input.}
    \label{fig:combined plot}
\end{figure}

As illustrated in Fig.~\ref{fig:combined plot}, the function consists of two major parts.
For regions where the uncertainty estimate $\psi_{est}(y_0^i) \in [0, \psi_{max}]$, we define $u'(\cdot)$ as a linear function with an offset $b_u$ and a slope of $(1-b_u)/\psi_{max}$, ensuring the output remains within the range $[b_u, 1]$.
This part comprises both low-uncertainty and high-uncertainty regions, and we find this linear modeling of the relationship between perturbation and uncertainty value offers a simple yet effective solution.
Meanwhile, the positive offset $b_u$ ensure a minimum noise level, preventing edge and texture areas from being assigned extremely low noise levels due to occasionally inaccurate uncertainty estimates.
We empirically find that setting $\psi_{max}=0.05$ and $b_u=0.4$ leads to better perceptual quality, and several experimental results are shown in Tab.~\ref{tab: ablation b_u}.
In contrast, for regions where the uncertainty estimate $\psi_{est}(y_0^i) \in (\psi_{max}, +\infty)$, we set their weighting coefficients to a constant, i.e., $w_u(y_0)=1.0$.
A large amount of isotropic noise is then applied in these areas to provide sufficient perturbation for the score estimation to ensure the perceptual quality.
In general, the weighting coefficient function $u'(\cdot)$ can be formulated as:

\begin{equation}
    u'(\psi) = 
    \begin{cases}
      \frac{(1-b_u)}{\psi_{max}}\psi + b_u & \text{ if } 0 \le \psi \le \psi_{max} \\
      1 & \text{ otherwise}
    \end{cases}.
\end{equation}

\begin{table}
\small
\setlength{\tabcolsep}{5pt}

\caption{Ablation study on effects of offset $b_u$ in the weighting coefficient function. The best results are highlighted in \textbf{bold}.}
\label{tab: ablation b_u}

\vspace{-3mm}
  \begin{center}
  \begin{tabular}{cc|ccc|ccc}
    \toprule
    \multirow{2}{*}{Model} &\multirow{2}{*}{$b_u$} & \multicolumn{3}{c|}{RealSR} & \multicolumn{3}{c}{RealSet} \\
    && CLIPIQA$\uparrow$ & MUSIQ$\uparrow$ & NIQE$\downarrow$ & CLIPIQA$\uparrow$ & MUSIQ$\uparrow$ & NIQE$\downarrow$ \\

    \midrule
    w/o offset & 0.0  & 0.4464 & 55.671 & 4.74 & 0.5445 & 57.335 & 4.84 \\
    w/$\ $   offset & 0.4  & \textbf{0.6010} & \textbf{64.541} & \textbf{4.02} & \textbf{0.6389} & \textbf{63.498} & \textbf{4.24} \\
    w/o uncertainty & 1.0  & 0.5191 & 61.728 & 4.40 & 0.5781 & 61.371 & 4.58 \\

    \bottomrule
  \end{tabular}
  \end{center}
\vspace{-5mm}
\end{table}

\section{Comparisons to pretraining-based SR methods.}
\label{sec:comparison1}
Pretraining-based SR methods~\cite{wang2024exploiting, xie2024addsr, wu2025one} harness the generative power from pretrained text-to-image models, such as stable diffusion, to enhance perceptual quality.
However, they follow an entirely different track from ours.
Firstly, their high perceptual quality comes from SD's capability to generate details inconsistent to LR inputs, therefore resulting in worse fidelity.
Secondly, the model size and GPU memory consumption of these methods surpass those of the proposed UPSR method by up to 10 times.
Thirdly, these methods are constrained by their fixed backbone, i.e., the stable diffusion model, making them less adaptable for rescaling. This limitation reduces their practicality when deployed on lightweight devices. 
Besides, these methods have to crop large input image to $128\times 128$ patches, while UPSR can be directly applied to $512\times 512$ or larger images.

\section{Comparisons to one-step methods.}
\label{sec:comparison2}
In the main paper, we apply five inference steps because we believe several more steps can better unleash the potential of diffusion models.
The performance of distillation-based methods (e.g., SinSR~\cite{wang2024sinsr}) is closely tied to that of their multi-step teacher models (e.g., ResShift~\cite{yue2024resshift}).
We therefore develop UPSR which could work as a better teacher model to support the training of better one-step models.
Meanwhile, the target of reducing diffusion steps is speeding up inference, which can also be achieved by rescaling the model size of denoiser.
To make comparison with SinSR, we present UPSR-light with smaller model size.
As shown in Tab.~\ref{tab:comp_sin}, the rescaled model achieves better performance with comparable inference speed and less than 1/3 of total model size.

\begin{table*}[h]
\scriptsize
\setlength{\tabcolsep}{2.5pt}

    \centering
    \begin{minipage}{0.49\textwidth}
        \centering
        \caption{Qualitative comparison with SD-based models on RealSR dataset. Quality metrics are re-evaluated on uncropped image.}
        \label{tab:comp_sd}
        \begin{tabular}{l|ccc|cccc}
            \toprule
             Models & Params & Runtime & Memory & PSNR$\uparrow$ & LPIPS$\downarrow$ & MUSIQ$\uparrow$ & NIQE$\downarrow$ \\
             \midrule
             StableSR-200 & 1410M & 33.0s & 8.51G & 25.80 &\textbf{ 0.2665} & 48.346 & 5.87 \\
             AddSR-1 & 2280M & 0.659s & 8.47G & 25.23 & 0.2986 & 63.011 & 5.17 \\
             OSEDiff-1 & 1775M & 0.310s & 5.85G & 24.57 & 0.3035 & \textbf{67.310} & 4.34 \\
             \textbf{UPSR}-5 & \textbf{122}M & \textbf{0.212}s & \textbf{2.83}G & \textbf{26.44} & 0.2871 & 64.541 & \textbf{4.02} \\
             \bottomrule
        \end{tabular}
        \label{tab:sdbased}
    \end{minipage}
    \hfill
    \begin{minipage}{0.49\textwidth}
        \centering
        \caption{Qualitative comparison with one-step model on RealSR dataset. Quality metrics are re-evaluated on uncropped image.}
        \label{tab:comp_sin}
        \begin{tabular}{l|ccc|cccc}
            \toprule
             Models & Params & Time & Memory & PSNR$\uparrow$ & LPIPS$\downarrow$ & MUSIQ$\uparrow$ & NIQE$\downarrow$ \\
             \midrule
             SinSR-1 & 174M & \textbf{0.141s} & 4.03G & 26.01 & 0.4015 & 59.344 & 6.26 \\
             \textbf{UPSR}-light-4 & \textbf{52.7M} & 0.148s & \textbf{2.48G} & \textbf{26.28} & \textbf{0.3025} & \textbf{63.785} & \textbf{4.13} \\
             \bottomrule
        \end{tabular}
        \label{tab:single}
    \end{minipage}
\end{table*}

\section{Additional Visual Examples}
\label{sec:more vis}
\begin{figure}
    \centering
    \includegraphics[width=\linewidth]{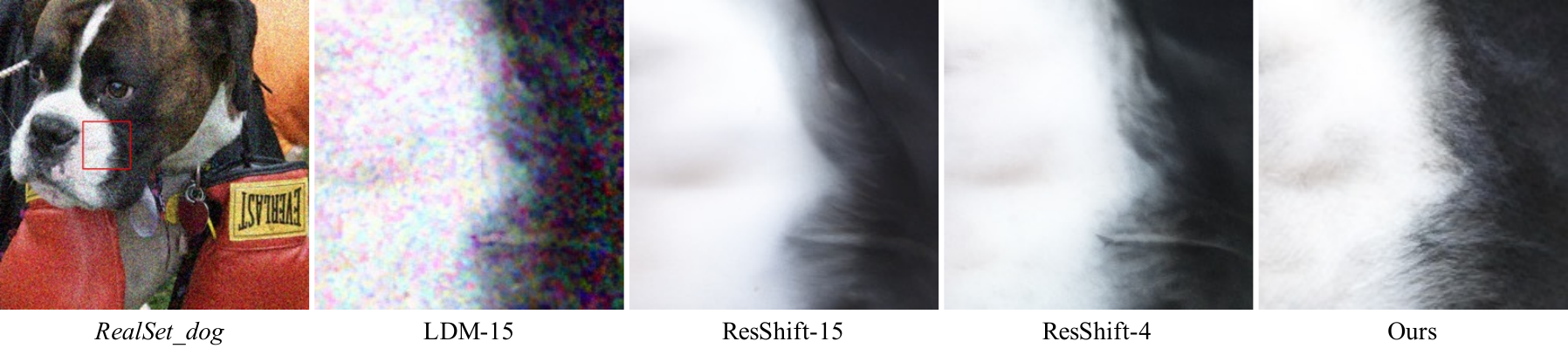}
    \includegraphics[width=\linewidth]{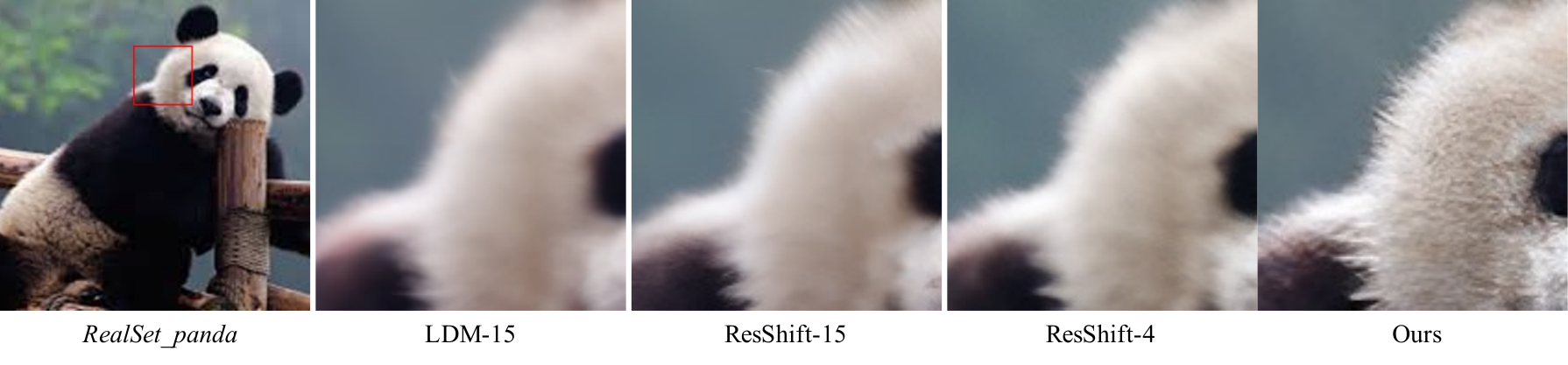}
    \caption{Additional visual comparisons on RealSet~\cite{yue2024resshift}.}
    \label{fig:additional vis realset}
\end{figure}

\begin{figure}[p]
    \centering
    \includegraphics[width=\linewidth]{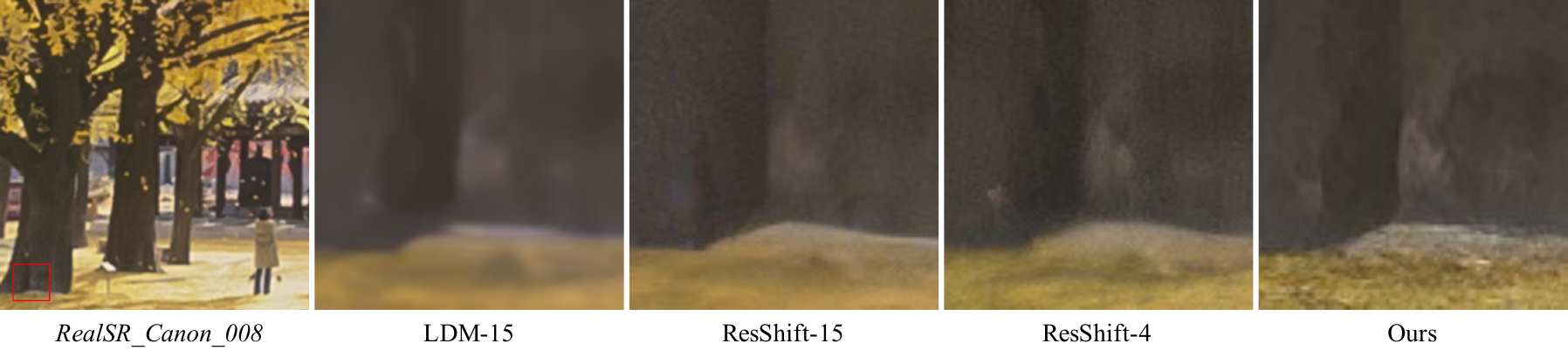}
    \includegraphics[width=\linewidth]{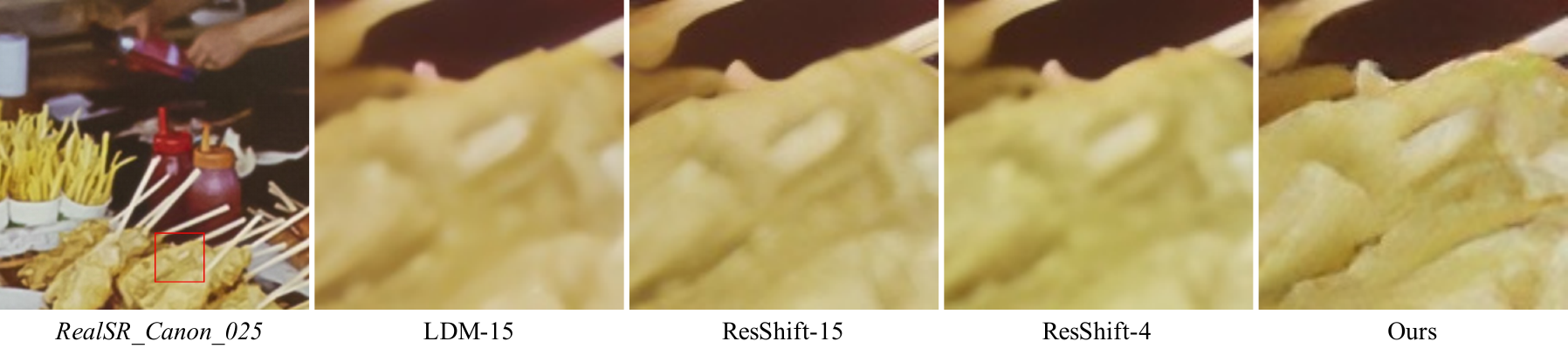}
    \includegraphics[width=\linewidth]{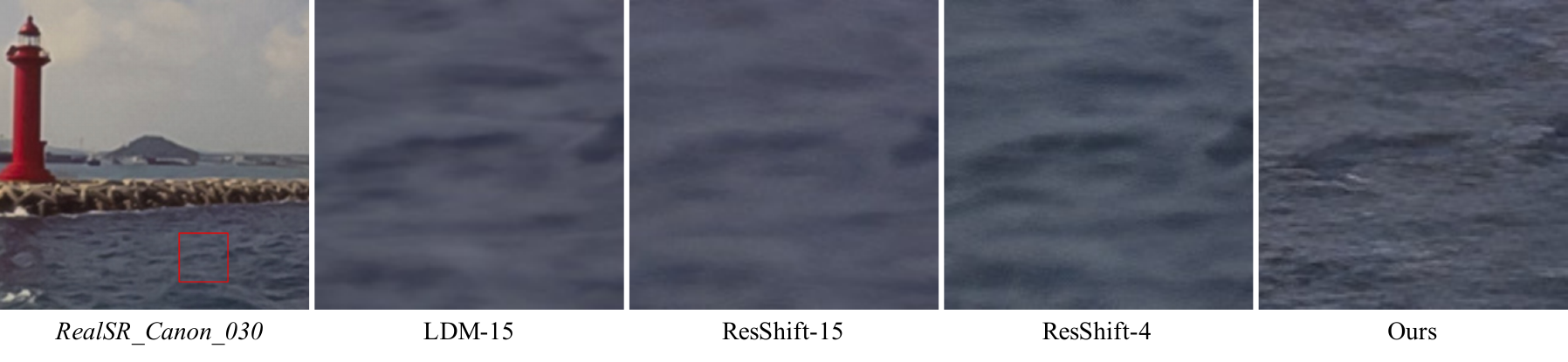}
    \includegraphics[width=\linewidth]{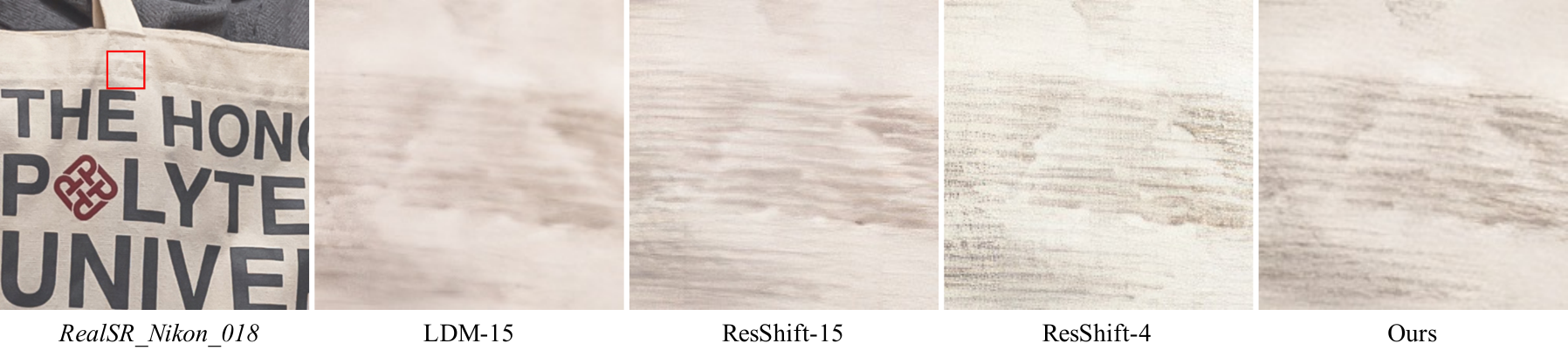}
    \caption{Additional visual comparisons on RealSR~\cite{cai2019toward}.}
    \label{fig:additional vis realsr}
\end{figure}

In Fig.~\ref{fig:additional vis realset} and Fig.~\ref{fig:additional vis realsr}, we present more visual comparisons between the proposed UPSR and existing diffusion-based SR methods~\cite{rombach2022high, yue2024resshift, yue2024efficient}.

\clearpage


\end{document}